\documentclass[10pt,twocolumn,letterpaper]{article}

\usepackage{amsmath,amssymb,amsthm}
\usepackage{graphicx}

\usepackage[left=2cm,top=2cm,right=2cm,bottom=2.5cm]{geometry}
\usepackage[compact]{titlesec}
\usepackage{abstract}
\usepackage{adjustbox}

\newcommand{\rot}[1]{\multicolumn{1}{c|}{\adjustbox{angle=60}{#1}}}

\usepackage[numbers,sectionbib]{natbib}

\usepackage{algorithm}
\usepackage{algorithmic}

\usepackage{xspace}
\newcommand{\trainA}{\textbf{VOC2012}\xspace}
\newcommand{\trainB}{\textbf{VOC2007+2012}\xspace}
\newcommand{\trainC}{\textbf{VOC+COCO}\xspace}

\usepackage{listings}
\usepackage{color}

\newcommand{\eg}{e.g.,~}
\newcommand{\ie}{i.e.,~}

\newcommand{\figref}[1]{Figure~\ref{#1}}

\def\thmcolon{\hspace{-.85em} {\bf :}}
\definecolor{bgCode}{rgb}{0.94, 0.94, 1.0}
\newcommand{\horizon}{\ensuremath{v^{\textrm{hz}}}}

\newcommand{\pixelSet}{{\cal P}}

\newcommand{\factor}[1]{\psi^{\textrm{#1}}}

\begin{document}

%%%%%%%%% TITLE
\title{Deep CNN Ensemble with Data
  Augmentation for Object Detection}

\author{Jian Guo\\
Research School of Computer Science\\
The Australian National University\\
{\tt\small jian.guo@anu.edu.au}
\and 
Stephen Gould\\
Research School of Computer Science\\
The Australian National University\\
{\tt\small stephen.gould@anu.edu.au}
}

\maketitle

% Abstract -------------------------------------------------------------------------

\begin{abstract}
We report on the methods used in our recent {\sc DeepEnsembleCoco}
submission to the PASCAL VOC 2012 challenge, which achieves
state-of-the-art performance on the object detection task. Our method
is a variant of the R-CNN model proposed by~\citet{Girshick:CVPR14}
with two key improvements to training and evaluation. First, our
method constructs an ensemble of deep CNN models with different
architectures that are complementary to each other. Second, we augment
the PASCAL VOC training set with images from the Microsoft COCO
dataset to significantly enlarge the amount training
data. Importantly, we select a subset of the Microsoft COCO images to
be consistent with the PASCAL VOC task. Results on the PASCAL VOC
evaluation server show that our proposed method outperform all
previous methods on the PASCAL VOC 2012 detection task at time of
submission.
\end{abstract}

% Introduction ---------------------------------------------------------------------
\section{Introduction}

In recent years, advances in deep learning have dramatically boosted
the performance of object recognition, detection and segmentation
tasks (\eg~see ~\cite{Krizhevsky:NIPS12}, ~\cite{Girshick:CVPR14} and
~\cite{Long:CVPR15}, respectively). Large-scale convolutional neural
networks (CNNs) pretrained on large datasets, such as
ImageNet~\cite{ImageNet:2009}, have demonstrated consistent
improvement and generalizability across other smaller datasets and all
current state-of-the-art results on the well known PASCAL VOC
dataset~\cite{Everingham:IJCV2010} use this approach. Thus, an
emergent trend in developing CNN models for computer vision
applications is to start from a pretrained neural network and then
fine tune the parameters for the task at hand, such as detection,
segmentation, or activity recognition, and specific domain (\ie
dataset).

In addition, current best practice suggests that combining the output
from several models and augmenting training data to improve the
variability of instances seen during learning are further ingredients
necessary for achieving state-of-the-art performance. Both of these
are well known techniques in the machine learning community and relate
to model averaging and over-fitting prevention. However, precise
details in their implementation can dramatically running times and
effectiveness.

In this technical report we detail our procedure for achieving
state-of-the-art performance on the PASCAL VOC detection
task. Different from existing methods, which use a single CNN model
fine tuned on the PASCAL VOC training set, we combine the practices
outlined above. Specifically, we construct an ensemble of CNN models
with different architectures with parameters learned on different
subsets of our augmented training set---a combination of the original
PASCAL VOC training set and the much larger Microsoft COCO
dataset~\cite{Lin:COCO}. We include experimental analysis on
components of our model and the final combined model that was
submitted to the PASCAL VOC evaluation server and achieved
state-of-the-art results at the time of submission (3 May
2015).\footnote{Our submission was subsequently beaten by the method
  of~\citet{Gidaris:arXiv2015} on 9 May 2015.}.

% Related Work ---------------------------------------------------------------------
\section{Related Work}

The introduction of the R-CNN approach by \citet{Girshick:CVPR14}
opened the door for features obtained through deep learning to improve
object detection performance on the PASCAL VOC dataset. In their work,
AlexNet CNN architecture~\cite{Krizhevsky:NIPS12} was used to extract
a set of deep features from arbitrary rectangular regions and used for
object classification. Since the introduction of AlexNet, deep
learning has advanced significantly both in terms of model
architecture and training methods. We hypothesize that improving the
feature extraction part of the pipeline by combining the recent
advances in deep learning can boost model performance. To this end,
our work replaces the single AlexNet model with an ensemble of
different models, namely GoogleNet~\cite{Szegedy:GoogleNet} and
VGG-16~\cite{Simonyan:ICLR2015}. These two recent models pushed the
error rate to below 10\% on ImageNet Large Scale Visual Recognition
Challenge (ILSVRC) 2014 competition~\cite{ILSVRC15}. In addition to
the improved network architectures, we also explore augmenting the
training dataset with images from the recently introduced Microsoft
COCO dataset~\cite{Lin:COCO}.

% Proposed Approach ---------------------------------------------------------------------

\section{Deep Ensemble Approach}

\begin{figure}[t]
\centering
\includegraphics[width=\columnwidth]{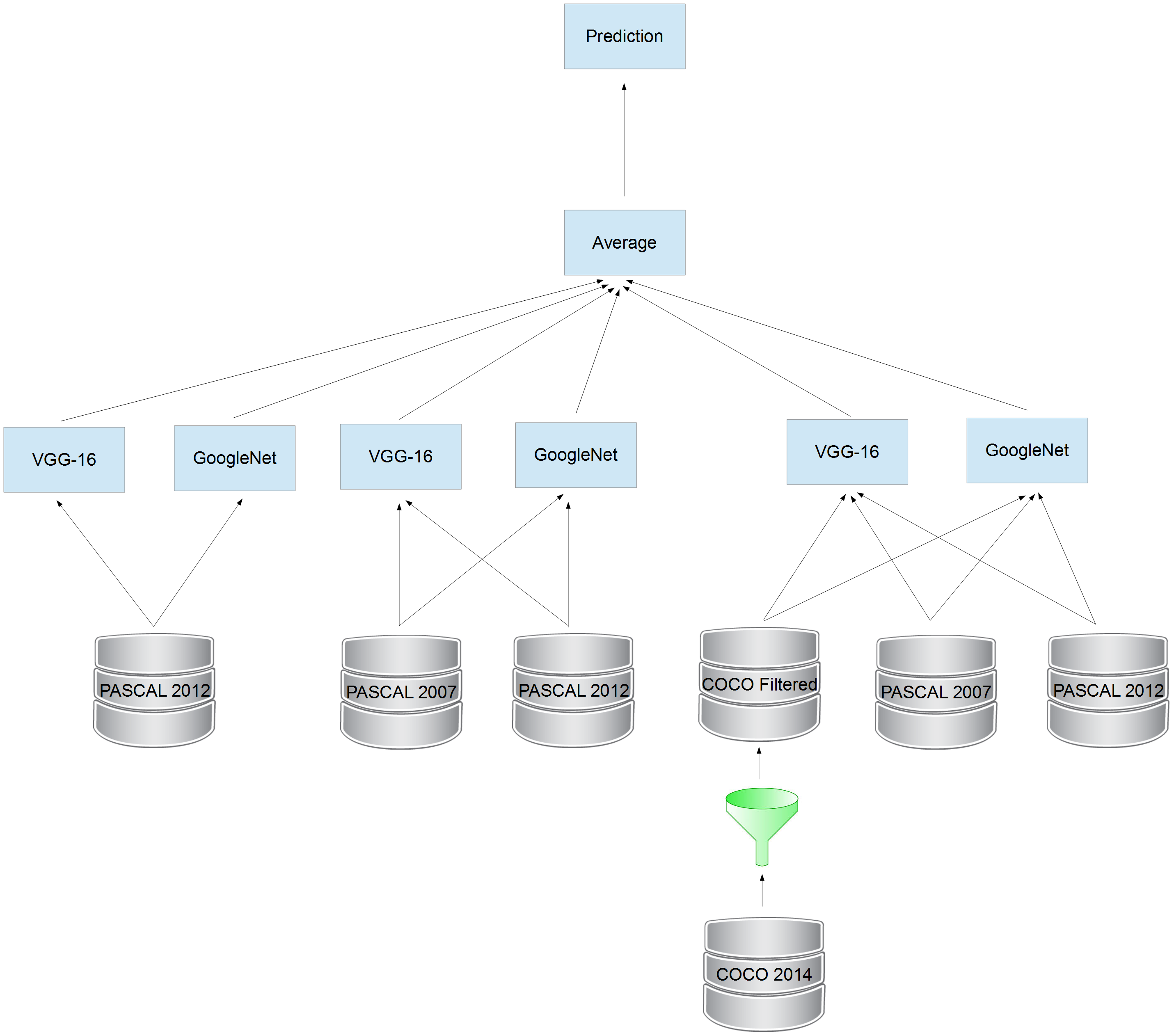}
\caption{Schematic of our deep ensemble model.}
\label{fig:model_illustration}
\end{figure}

We proposed an improved variant of the Region CNN (R-CNN) method of
\citet{Girshick:CVPR14} for better object detection. The improvement
comes from three well known but essential machine learning practices:
starting from good initial parameters, averaging models, and using as
much data as possible for training. An illustration of our method is
shown in Figure~\ref{fig:model_illustration} and overview as follows:

\begin{itemize}
  \item We start from two state-of-the-art pretrained networks,
    namely, GoogleNet~\cite{Szegedy:GoogleNet} and
    VGG-16~\cite{Simonyan:ICLR2015}. In contrast, the original R-CNN
    method starts from the pretrained AlexNet
    network~\cite{Krizhevsky:NIPS12}.
  \item We next refine the network parameters using combinations of
    existing datasets in three different ways. First, using the PASCAL
    VOC 2012 \texttt{train} set~\cite{Everingham:IJCV2010}. Second,
    using a merged training set consisting of all images from PASCAL
    VOC 2007 \texttt{trainval} and PASCAL VOC 2012 \texttt{train}. And
    third, using the above augmented with Microsoft COCO 2014
    \texttt{trainval}~\cite{Lin:COCO}. In the following we refer to
    these training sets as \trainA, \trainB, and \trainC,
    respectively.
  \item We finally combine the output of the refined GoolgeNet and
    VGG-16 networks by averaging their predictions. The averaged
    predictions outperform predictions from either network
    alone. Thus, we hypothesize that GoogleNet and VGG-16 learn
    complementary features.
\end{itemize}

\subsection{Training Baseline Models}

In our work we use a heterogeneous GPU cluster for training and
evaluation. We fine tune our baseline models on \trainA using Caffe
~\cite{jia2014caffe} and NVIDIA K20 GPUs and follow the protocol
detailed in~\citet{Girshick:CVPR14}. That is, we use stochastic
gradient descent (SGD) with initial learning rate of $10^{-3}$ and
decrease by a factor of 0.1 every 20,000 iterations. We also use
momentum of 0.9 and weight decay of $5 \times 10^{-4}$. We train for a
total of 100,000 iterations. Due to large memory requirements for
GoogleNet and VGG-16 (compared to AlexNet), we use different minibatch
sizes from the original R-CNN setup. Specifically, we use minibatch
sizes of 64 and 20 for GoogleNet and VGG-16, respectively. We also
omit the two small auxiliary GoogleNet convolutional networks during
fine tuning due to memory limitation. That is, we delete the
\texttt{loss1} and \texttt{loss2} branches from GoogleNet
network. This same fine tuning setup was used on all our models with
the set of training images changed accordingly.

After fine tuning both networks, we extract feature vectors for object
classification. From the GoogleNet network, we extract 1024 features
from the output of the last average pooling layer (i.e.,~immediately
before the 1024-dimensional fully connected layer). From VGG-16, we
extract the 4096-dimensional output of the first fully connected layer
after the rectified linear units (ReLU).

Using these 1024-dimensional and 4096-dimensional feature vectors from
GoogleNet and VGG-16, respectively, we train separate linear SVM
classifiers for each class independently. Here we use negative mining
and run the same post-processing pipeline as detailed
in~\citet{Girshick:CVPR14}. In addition, we also include experiment results
with bounding box regression.

\subsection{Combining GoogleNet and VGG-16}

There are many strategies that can be used to combine the output from
different models. For example, one could concatenate the feature
vectors from the different models and train a single classifier over
the higher dimensional input. Another approach is to compute a
straightforward average of the outputs from the models.

In early experiments we found that there was negligible difference in
accuracy between these two strategies. As such, we report results
using the simpler strategy of training the GoogleNet and VGG-16
networks separately and averaging their predictions at test time.

\subsection{Data Augmentation}

During informal testing we observed a large gap between performance on
the \texttt{train} and \texttt{val} datasets, the latter not used for
estimation of the model parameters. A natural conclusion then is that
the fine tuning process is overfitting to the training set. To combat
overfitting we augment the PASCAL VOC \texttt{train} dataset with
additional images, which we source in two different ways. Our
experimental results demonstrate the effectiveness of this strategy in
reducing the gap between the mean average precision on the
\texttt{train} and \texttt{val} datasets.

Our first approach is to merge the PASCAL VOC 2012 \texttt{train} set
(which we call \trainA) containing 5,717 labelled images with the
PASCAL VOC 2007 \texttt{trainval} set containing 5,011 labelled
images. This produces a training set that is almost double in size. We
call the new training set \trainB. Since the two datasets being merged
share the same class labels combining them is straightforward.

Our second approach to data augmentation is not as simple as the
first. Here we combine \trainB with data from the recently released
Microsoft COCO \texttt{trainval} dataset~\cite{Lin:COCO}. We call the
resulting dataset \trainC. However, in order to produce this dataset
we need to overcome three challenges. First, the Microsoft COCO
dataset consists of many small objects, much smaller than the objects
annotated in the PASCAL VOC datasets.\footnote{By small we mean that
  the object's ground truth bounding box has width or height less than
  30 pixels.} We hypothesize that these objects will be problematic
for training a model that will not encounter such small objects at
test time. As such, we simply filter them out prior to merging.

The second challenge we need to overcome is that the Microsoft COCO
dataset annotates objects with a different set of categories to the
labels used in PASCAL VOC datasets. Microsoft COCO has eighty
categories while PASCAL VOC has only twenty. Nevertheless, many of the
Microsoft COCO categories can be mapped onto the PASCAL VOC
classes. For example, the \emph{couch} label in COCO corresponds to
the \emph{sofa} label in PASCAL. Here we fine tune the CNN model
parameters using all eighty COCO classes, with the PASCAL VOC classes
mapped to corresponding classes. The final SVM classifiers are then
trained on the twenty PASCAL VOC classes (and only the PASCAL VOC
data). See Appendix~\ref{app:mapping} for the mapping used.

The third challenge to overcome is the practical memory limitations we
face when dealing with such large
datasets. In~\citet{Girshick:CVPR14}, selective
search~\cite{Uijlings:IJCV13} was used to generate approximately 2000
candidate bounding boxes per image. This already gives a very large
number of training examples for \trainB and therefore large memory
(i.e., disk) and processing requirements. We cannot currently
accommodate the massive increase in resources that would be required if
the same procedure was adopted for the Microsoft COCO data. Thus,
rather than use selective search for generating training data from the
Microsoft COCO dataset, we keep only the ground truth bounding boxes
(i.e., positive examples) and randomly select a small number of
negative examples from each image. We sample three negative examples
per ground example and having no overlap with any ground truth
bounding box within the image. This approach has the effect of
increasing the ratio of positive to negative training examples, which
are already well represented in \trainB.

Note that we only use \trainC for fine tuning of the GoogleNet and
VGG-16 network parameters. For training the final SVM classifiers, we
discard training examples from the Microsoft COCO dataset that do not
correspond to any of the twenty PASCAL VOC categories. The effective size
of the resulting training set is 105,815 images, almost ten times
larger than \trainB.

With this larger training set we fine tune the parameters on an NVIDIA
K80 GPU and increase the minibatch size to 128 and 82 for GoogleNet
and VGG-16, respectively.

% Experiments ---------------------------------------------------------------------

\section{Experiments and Results}

In this section we evaluate our proposed training methods on the
PASCAL VOC 2012 validation set. We further report results obtained on
the PASCAL VOC 2012 by submitting a model to the PASCAL evaluation
server. Here we additionally fine tune our model on the validation set
images.

\subsection{Baseline Models (\trainA)}

As can be seen from the results in
Table~\ref{table:tbl_voc_2012_results}, GoogleNet and VGG-16 trained
on \trainA give 59.4\% and 58.6\% mAP, respectively, on PASCAL VOC
2012 validation set. Once combined, the performance is boosted to
63.7\%. This suggests that the two networks learn complementary
features such that one tends to correct the other ones mistakes.

\subsection{Data Augmentation}

In these experiments we evaluate the affect of enlarging the training
set via data augmentation. Here we merge the PASCAL VOC 2007 train and
validation sets with the PASCAL VOC 2012 train set and fine tune our
parameters on this combined set.

As can be seen from the results in
Table~\ref{table:tbl_voc_2012_results}, GoogleNet and VGG-16 trained
on \trainB give 62.1\% and 60.5\% mAP, respectively. This represents
about 2\% improvement over the baseline models. The combined
performance is 65.0\%, which is 1.3\% better than the combined
baseline. Thus we can see that performance is consistently improved
for both of the networks independently as well as the combined,
validating the intuition that more (labeled) training data helps fine
tuning. Note, however, that the improvement gain when combining the
models trained on a larger dataset is less than the improvement gain
when combining the baseline models. This suggests a diminishing return
on performance as more data is used for training.

\subsection{Combining Four Networks}

Next, we evaluate performance when combining four networks---GoogleNet
and VGG-16 trained with and without data augmentation. As can be seen
in Table~\ref{table:tbl_voc_2012_results}, the combination of four
networks results in 66.0\% mAP, which is about 1\% improvement over
the previous two networks combined. Thus there is still value in
including the baseline models in our ensemble average to provide
complementary information and reduce any affect of overfitting.

\subsection{Further Data Augmentation}

In addition to the four network mentioned above, we evaluate
performed on two additional networks trained using the Microsoft COCO
data augmentation strategy. This gives us six models, the combination
of which results in 68.3\% mAP on the PASCAL VOC 2012 validation set
(Table~\ref{table:tbl_voc_2012_results}). This represents a 2.3\%
improvement over the combined previous four networks.

\subsection{Bounding Box Regression and Averaging}

Following the approach of~\citet{Girshick:CVPR14}, we applied bounding
box regression to the predictions for each of the trained
networks. The selective search procedure proposes 2,000 bounding boxes
per image, which results in 12,000 regressed boxes once we apply
bounding box regression to each of our six networks. We average the
bounding boxes by feeding the average SVM score across the six
networks for each selective search box and averaging the four
regressed coordinates. This results in a further performance
improvement of 2\%.

\subsection{PASCAL VOC Test Server Results}

To assess our performance on completely unseen data we prepared a
submission to the PASCAL VOC evaluation server. Here we used the
procedure same as above with the addition of two more networks
(GoogleNet and VGG-16) fine tuned on \trainC augmented with the PASCAL
VOC 2012 validation set images. In addition the final SVM classifiers
were trained using both training and validation sets.

Our test results can be seen in
Table~\ref{table:tbl_voc_2012_test_results} and example results on a
handful of categories in \figref{fig:examples}. Our model was the top
ranked solution at the time of submission (3 May 2015). A subsequent
submission~\cite{Gidaris:arXiv2015} outperforms our model by 0.6\%
and is included in Table~\ref{table:tbl_voc_2012_test_results} for
reference.

% Conclusion ---------------------------------------------------------------------

\section{Conclusion}

This paper describes our submission on 3 May 2015 to the PASCAL VOC
test server for the object detection challenge. Our work confirms two
important best practices used in the training of machine learning
models. First, that fine tuning performance can be improved with more
training data. Second, that the overall accuracy is increased when
averaging the output of models trained on different datasets (or
components of datasets). As the quantity of training data increases,
however, the performance improvement of the ensemble diminishes. These
simple techniques, while not new, allowed us to achieve
state-of-the-art performance on the PASCAL VOC object detection
challenge.

%------- tables ---------

\begin{center}
\begin{table*}
    \scriptsize
    \centering%
    \setlength{\tabcolsep}{2pt}
    \begin{tabular}{ | l | l | l | l | l | l | l | l | l | l | l | l | l | l | l | l | l | l | l | l | l | l | p{1cm} |}
    \hline
    VOC 2012 Val & \rot{aero} & \rot{bike} & \rot{bird} & \rot{boat} & \rot{bottle} & \rot{bus} & \rot{car} & \rot{cat} & \rot{chair} & \rot{cow} & \rot{table} & \rot{dog} & \rot{horse} & \rot{motor} & \rot{person} & \rot{plant} & \rot{sheep} & \rot{sofa} & \rot{train} & \rot{tv} & mAP \\ \hline
    \multicolumn{22}{|c|}{GoogleNet and VGG-16 Trained on PASCAL 2012}\\ \hline
    GoogleNet & 74.1 & 68.9 & 59.9 & 36.7 & 35.4 & 71.6 & 62.4 & 81.8 & 36.1 & 58.5 & 40.0 & 77.5 & 67.8 & 74.8 & 61.0 & 30.8 & 61.2 & 58.0 & 67.0 & 64.1 & 59.4 \\ \hline    
    VGG-16 & 73.6 & 69.7 & 55.3 & 35.6 & 33.5 & 72.5 & 60.3 & 80.1 & 34.6 & 57.7 & 42.5 & 76.2 & 68.2 & 75.1 & 60.0 & 30.0 & 61.5 & 56.3 & 64.8 & 63.5 & 58.6 \\ \hline
    \multicolumn{22}{|c|}{GoogleNet and VGG-16 Average Trained on PASCAL VOC 2012}\\ \hline 
    GoogleNet,Vgg-16 & 78.6 & 73.7 & 63.9 & 40.8 & 39.6 & 74.8 & 65.2 & 85.4 & 40.5 & 65.5 & 47.6 & 81.8 & 72.8 & 78.3 & 63.5 & 35.8 & 65.6 & 62.7 & 70.3 & 68.0 & 63.7 \\ \hline
    \multicolumn{22}{|c|}{GoogleNet and VGG-16 Trained on PASCAL VOC 2007 \& 2012}\\ \hline
    GoogleNet & 74.3 & 70.5 & 63.7 & 40.8 & 38.5 & 74.6 & 64.2 & 86.1 & 35.9 & 61.7 & 41.4 & 80.2 & 72.7 & 77.1 & 62.7 & 35.2 & 63.7 & 58.8 & 71.0 & 68.5 & 62.1 \\ \hline    
    VGG-16 & 73.9 & 71.8 & 57.5 & 37.2 & 35.3 & 73.1 & 62.9 & 83.2 & 36.5 & 61.2 & 45.4 & 79.1 & 70.0 & 76.8 & 61.7 & 32.8 & 62.9 & 58.3 & 66.8 & 63.9 & 60.5 \\ \hline
    \multicolumn{22}{|c|}{GoogleNet and VGG-16 Average Trained on PASCAL VOC 2007 \& 2012}\\ \hline
    GoogleNet,Vgg-16 & 76.4 & 74.1 & 66.1 & 43.7 & 42.5 & 76.8 & 66.7 & 87.1 & 39.5 & 65.0 & 48.2 & 83.3 & 74.7 & 79.4 & 65.6 & 37.6 & 66.5 & 63.6 & 73.5 & 70.1 & 65.0 \\ \hline
    \multicolumn{22}{|c|}{4 Nets Average}\\ \hline
    4 nets avg & 78.2 & 75.2 & 66.6 & 44.2 & 42.4 & 76.9 & 67.0 & 87.3 & 42.3 & 66.7 & 51.2 & 84.4 & 74.5 & 80.3 & 65.8 & 39.9 & 66.8 & 65.9 & 73.4 & 70.0 & 66.0 \\ \hline
    \multicolumn{22}{|c|}{6 Nets Average}\\ \hline
    6 nets avg & 79.8 & 76.5 & 68.5 & 47.9 & 45.4 & 78.6 & 69.0 & 88.4 & 46.4 & 69.3 & 53.6 & 85.6 & 77.7 & 81.0 & 67.5 & 42.9 & 69.1 & 69.6 & 75.9 & 72.7 & 68.3 \\ \hline
    \multicolumn{22}{|c|}{6 Nets Average (bbox reg)}\\ \hline
     6 nets(bbox reg) & \textbf{82.2} & \textbf{78.9} & \textbf{72.1} & \textbf{51.6} & \textbf{49.9} & \textbf{79.0} & \textbf{70.9} & \textbf{89.6} & \textbf{48.3} & \textbf{69.7} & \textbf{53.9} & \textbf{87.4} & \textbf{79.3} & \textbf{82.2} & \textbf{70.6} & \textbf{45.7} & \textbf{71.2} & \textbf{71.0} & \textbf{77.5} & \textbf{74.6} & \textbf{70.3} \\ \hline
     
    \end{tabular}
    \caption{GoogleNet and VGG-16 on PASCAL VOC 2012 validation set.}\label{table:tbl_voc_2012_results} 
    \end{table*}
\end{center}

\begin{center}
\begin{table*}
    \scriptsize
    \centering%
    \setlength{\tabcolsep}{1.9pt}
    \begin{tabular}{ | l | l | l | l | l | l | l | l | l | l | l | l | l | l | l | l | l | l | l | l | l | l | p{1cm} |}
    \hline
    VOC 2012 Test & \rot{aero} & \rot{bike} & \rot{bird} & \rot{boat} & \rot{bottle} & \rot{bus} & \rot{car} & \rot{cat} & \rot{chair} & \rot{cow} & \rot{table} & \rot{dog} & \rot{horse} & \rot{motor} & \rot{person} & \rot{plant} & \rot{sheep} & \rot{sofa} & \rot{train} & \rot{tv} & mAP \\ \hline
    8 nets(bbox reg) & 84.0 & 79.4 & \textbf{71.6} & \textbf{51.9} & 51.1 & \textbf{74.1} & 72.1 & \textbf{88.6} & 48.3 & 73.4 & 57.8 & \textbf{86.1} & \textbf{80.0} & 80.7 & 70.4 & \textbf{46.6} & \textbf{69.6} & \textbf{68.8} & 75.9 & 71.4 & 70.1 \\ \hline
    State of the art~\cite{Gidaris:arXiv2015} & \textbf{85.0} & \textbf{79.6} & 71.5 & 55.3 & \textbf{57.7} & 76.0 & \textbf{73.9} & 84.6 & \textbf{50.5} & \textbf{74.3} & \textbf{61.7} & 85.5 & 79.9 & \textbf{81.7} & \textbf{76.4} & 41.0 & 69.0 & 61.2 & \textbf{77.7} & \textbf{72.1} & \textbf{70.7} \\ \hline 
    \end{tabular}
    \caption{PASCAL VOC 2012 test set results.}\label{table:tbl_voc_2012_test_results} 
    \end{table*}
\end{center}

\begin{figure}[t]
\centering
\setlength{\tabcolsep}{2pt}
\begin{tabular}{c}
  \includegraphics[height=3.875cm]{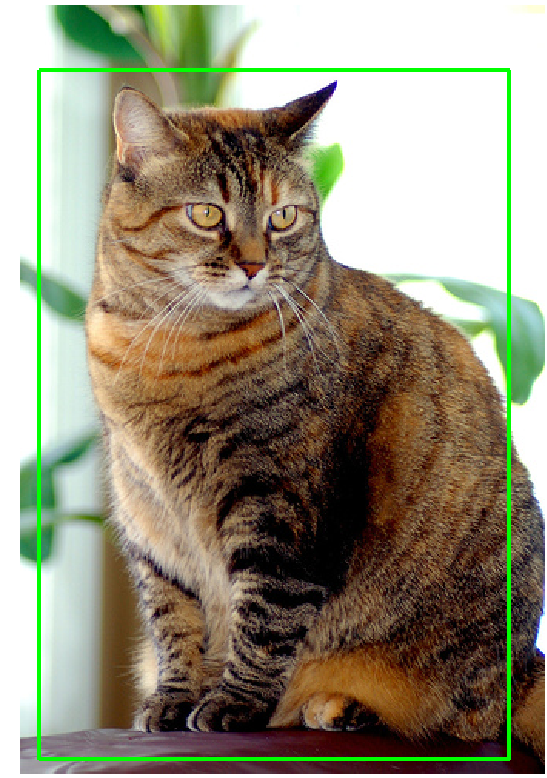}
  \includegraphics[height=3.875cm]{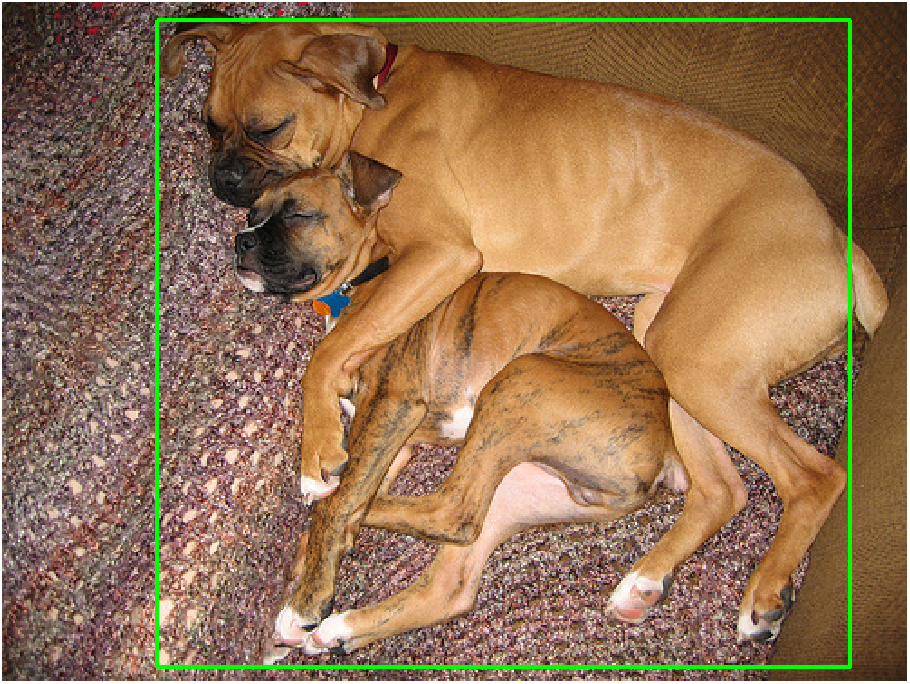} \\
  \includegraphics[height=3.65cm]{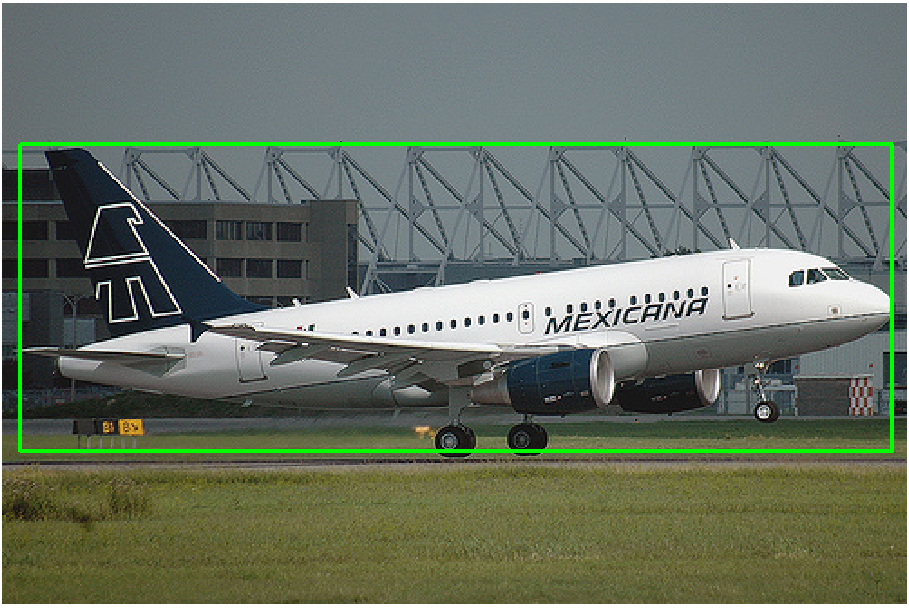}
  \includegraphics[height=3.65cm]{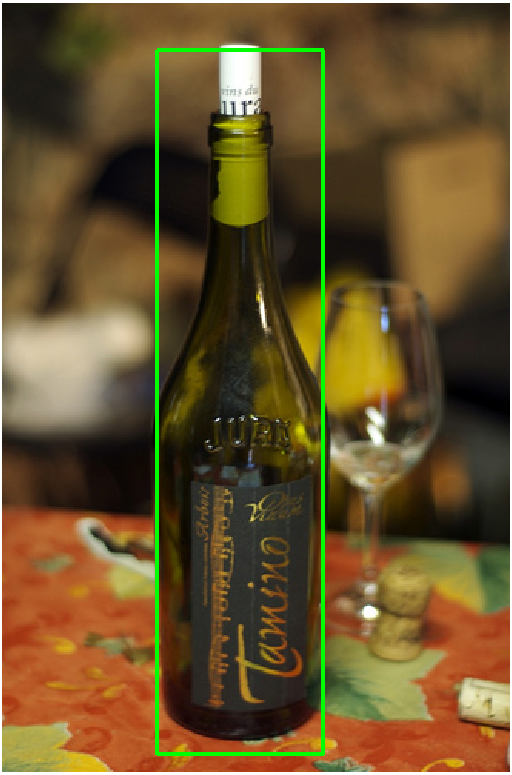} \\
  \includegraphics[height=3.5cm]{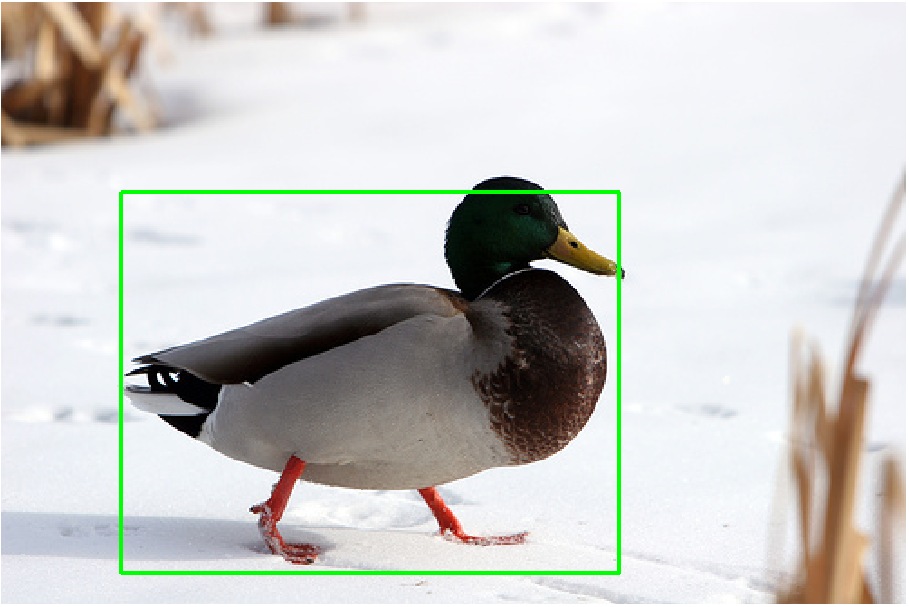}
  \includegraphics[height=3.5cm]{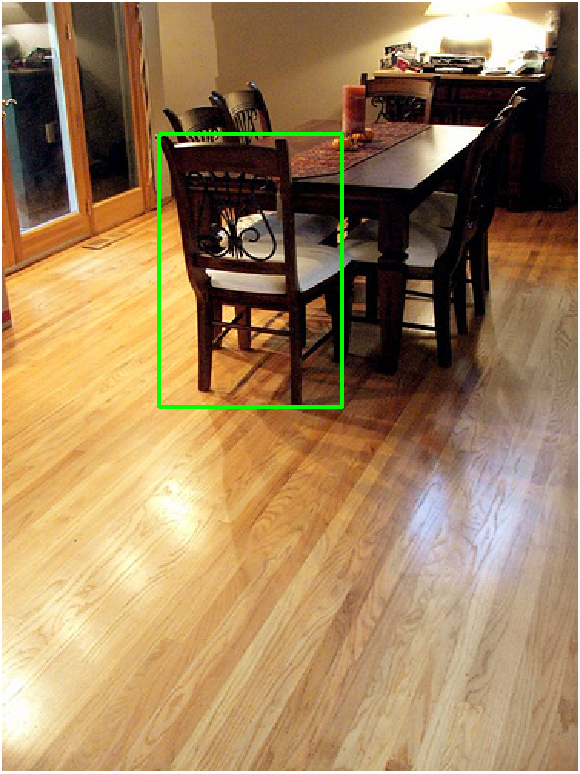} \\
\end{tabular}
\caption{\label{fig:examples} Some example detections on various
  categories correctly classified by our deep ensemble model.}
\end{figure}

% Bibliography --------------------------------------------------------------------
% can be 9-th page
{
  \small
  \setlength{\bibsep}{0mm}
  \bibliographystyle{plainnat}

\begin{thebibliography}{12}
\providecommand{\natexlab}[1]{#1}
\providecommand{\url}[1]{\texttt{#1}}
\expandafter\ifx\csname urlstyle\endcsname\relax
  \providecommand{\doi}[1]{doi: #1}\else
  \providecommand{\doi}{doi: \begingroup \urlstyle{rm}\Url}\fi

\bibitem[Deng et~al.(2009)Deng, Dong, Socher, Li, Li, and
  Fei-Fei]{ImageNet:2009}
J.~Deng, W.~Dong, R.~Socher, L.-J. Li, K.~Li, and L.~Fei-Fei.
\newblock Image{N}et: A large-scale hierarchical image database.
\newblock 2009.

\bibitem[Everingham et~al.(2010)Everingham, Van~Gool, Williams, Winn, and
  Zisserman]{Everingham:IJCV2010}
M.~Everingham, L.~Van~Gool, C.~K.~I. Williams, J.~Winn, and A.~Zisserman.
\newblock The pascal visual object classes ({VOC}) challenge.
\newblock 88\penalty0 (2):\penalty0 303--338, June 2010.

\bibitem[Gidaris and Komodakis(2015)]{Gidaris:arXiv2015}
Spyros Gidaris and Nikos Komodakis.
\newblock Object detection via a multi-region {\&} semantic segmentation-aware
  {CNN} model.
\newblock \emph{arXiv preprint arXiv:1505.01749}, 2015.

\bibitem[Girshick et~al.(2014)Girshick, Donahue, Darrell, and
  Malik]{Girshick:CVPR14}
R.~Girshick, J.~Donahue, T.~Darrell, and J.~Malik.
\newblock Rich feature hierarchies for accurate object detection and semantic
  segmentation.
\newblock 2014.

\bibitem[Jia et~al.(2014)Jia, Shelhamer, Donahue, Karayev, Long, Girshick,
  Guadarrama, and Darrell]{jia2014caffe}
Yangqing Jia, Evan Shelhamer, Jeff Donahue, Sergey Karayev, Jonathan Long, Ross
  Girshick, Sergio Guadarrama, and Trevor Darrell.
\newblock Caffe: Convolutional architecture for fast feature embedding.
\newblock \emph{arXiv preprint arXiv:1408.5093}, 2014.

\bibitem[Krizhevsky et~al.(2012)Krizhevsky, Sutskever, and
  Hinton]{Krizhevsky:NIPS12}
A.~Krizhevsky, I.~Sutskever, and G.~E. Hinton.
\newblock {ImageNet} classification with deep convolutional neural networks.
\newblock 2012.

\bibitem[Lin et~al.(2014)Lin, Maire, Belongie, Bourdev, Girshick, Hays, Perona,
  Ramanan, Zitnick, and Dollár]{Lin:COCO}
Tsung-Yi Lin, Michael Maire, Serge Belongie, Lubomir Bourdev, Ross Girshick,
  James Hays, Pietro Perona, Deva Ramanan, C.~Lawrence Zitnick, and Piotr
  Dollár.
\newblock Microsoft {COCO}: Common objects in context.
\newblock 2014.

\bibitem[Long et~al.(2015)Long, Shelhamer, and Darrell]{Long:CVPR15}
Jonathan Long, Evan Shelhamer, and Trevor Darrell.
\newblock Fully convolutional networks for semantic segmentation.
\newblock 2015.

\bibitem[Russakovsky et~al.(2015)Russakovsky, Deng, Su, Krause, Satheesh, Ma,
  Huang, Karpathy, Khosla, Bernstein, Berg, and Fei-Fei]{ILSVRC15}
O.~Russakovsky, J.~Deng, H.~Su, J.~Krause, S.~Satheesh, S.~Ma, Z.~Huang,
  A.~Karpathy, A.~Khosla, M.~Bernstein, A.C. Berg, and Li~Fei-Fei.
\newblock {ImageNet} large scale visual recognition challenge.
\newblock 2015.

\bibitem[Simonyan and Zisserman(2015)]{Simonyan:ICLR2015}
Karen Simonyan and Andrew Zisserman.
\newblock Very deep convolutional networks for large-scale image recognition.
\newblock In \emph{ICLR}, 2015.

\bibitem[Szegedy et~al.(2015)Szegedy, Liu, Jia, Sermanet, Reed, Anguelov,
  Erhan, Vanhoucke, and Rabinovich]{Szegedy:GoogleNet}
Christian Szegedy, Wei Liu, Yangqing Jia, Pierre Sermanet, Scott Reed, Dragomir
  Anguelov, Dumitru Erhan, Vincent Vanhoucke, and Andrew Rabinovich.
\newblock Going deeper with convolutions.
\newblock 2015.

\bibitem[Uijlings et~al.(2013)Uijlings, van~de Sande, Gevers, and
  Smeulders]{Uijlings:IJCV13}
J.R.R. Uijlings, K.E.A. van~de Sande, T.~Gevers, and A.W.M. Smeulders.
\newblock Selective search for object recognition.
\newblock 2013.

\end{thebibliography}

}

% Appendix ------------------------------------------------------------------------

\appendix

\section{PASCAL VOC to Microsoft COCO Class Mapping}
\label{app:mapping}

\begin{table}[h!]
    \centering%
    \begin{tabular}{ | l | l |}
    \hline
    PASCAL VOC & COCO \\ \hline
    \textbf{aeroplane} & \textbf{airplane} \\ \hline
    \textbf{bike} & \textbf{bicycle} \\ \hline
    bird & bird \\ \hline
    boat & boat \\ \hline
    bottle & bottle \\ \hline
    bus & bus \\ \hline
    car & car \\ \hline
    cat & cat \\ \hline
    chair & chair \\ \hline
    cow & cow \\ \hline
    dining table & dining table \\ \hline
    dog & dog \\ \hline
    horse & horse \\ \hline
    \textbf{motorbike} & \textbf{motorcycle} \\ \hline
    person & person \\ \hline
    potted plant & potted plant \\ \hline
    sheep & sheep \\ \hline
    \textbf{sofa} & \textbf{couch} \\ \hline
    train & train \\ \hline
    tv & tv \\ \hline
    \end{tabular}
    \caption{Mapping between PASCAL VOC and Microsoft COCO classes.}\label{table:tbl_pascal_coco} 
\end{table}

\end{document}